\documentclass{article}

\usepackage{arxiv}

\usepackage[utf8]{inputenc} 
\usepackage[T1]{fontenc}    
\usepackage{hyperref}       
\usepackage{url}            
\usepackage{booktabs}       
\usepackage{amsfonts}       
\usepackage{nicefrac}       
\usepackage{microtype}      
\usepackage{graphicx}
\usepackage{natbib}
\usepackage{doi}
\usepackage{algorithm}
\usepackage{amsmath}
\usepackage{algpseudocode}
\usepackage{float} 
\usepackage{amsmath,amssymb} 

\newcommand\blfootnote[1]{}
\title{DeepHive: A multi-agent reinforcement learning approach for automated discovery of swarm-based optimization policies}

\author{ \href{https://orcid.org/0000-0002-6438-1389}{\includegraphics[scale=0.06]{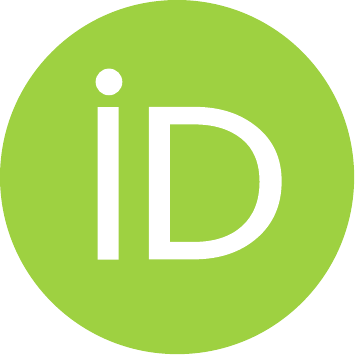}\hspace{1mm}Eloghosa A.~Ikponmwoba}\\
	Department of Mechanical Engineering\\
	Louisiana State University\\
	Baton Rouge, LA 70803 \\
	\texttt{eikpon1@lsu.edu} \\
	\And
	\href{https://orcid.org/0000-0001-5531-2245}{\includegraphics[scale=0.06]{orcid.pdf}\hspace{1mm}Opeoluwa ~Owoyele} \\
	Department of Mechanical Engineering\\
	Louisiana State University\\
	Baton Rouge, LA 70803 \\
	\texttt{oowoyele@lsu.edu} \\
}

\renewcommand{\shorttitle}

\hypersetup{
pdftitle={A template for the arxiv style},
pdfsubject={q-bio.NC, q-bio.QM},
pdfauthor={David S.~Hippocampus, Elias D.~Striatum},
pdfkeywords={First keyword, Second keyword, More},
}

\begin{document}
\maketitle

\begin{abstract}
We present an approach for designing swarm-based optimizers for the global optimization of expensive black-box functions. In the proposed approach, the problem of finding efficient optimizers is framed as a reinforcement learning problem, where the goal is to find optimization policies that require a few function evaluations to converge to the global optimum. The state of each agent within the swarm is defined as its current position and function value within a design space and the agents learn to take favorable actions that maximize reward, which is based on the final value of the objective function. The proposed approach is tested on various benchmark optimization functions and compared to the performance of other global optimization strategies. Furthermore, the effect of changing the number of agents, as well as the generalization capabilities of the trained agents are investigated. The results show superior performance compared to the other optimizers, desired scaling when the number of agents is varied, and acceptable performance even when applied to unseen functions. On a broader scale, the results show promise for the rapid development of domain-specific optimizers
\end{abstract}


\section{Introduction}
\par Design optimization, which involves the selection of design variables to achieve desired objectives, is important to many areas of engineering.  Manufacturing processes must be optimized to reduce waste and power consumption while simultaneously maximizing desirable outcomes such as quality control and manufacturing speed. The discovery of new materials with superior mechanical, electrical, or thermal properties can also be framed as an optimization problem, where the goal is to find the processing and compositional parameters that maximize the desired properties. Other examples include the optimization of lithium-ion batteries to reduce the risk of thermal runaway, minimizing drag on airfoil or wind turbine blades, and optimizing the combustion processes in an engine to reduce greenhouse gas emissions, just to name a few. In some of these examples, the process or performance of the system may be affected by a large number of variables interacting in complicated ways \citet{liao2010two}. In some cases, the problem of optimization is further complicated by the presence of several local optima, an objective function that is expensive or time-consuming to evaluate, or the existence of constraints. \blfootnote{Preprint submitted to Applied Soft Computing}

One way to classify existing algorithms for design optimization is by their utilization of derivatives for finding optimal designs. On the one hand, gradient-based optimizers use local information of the function’s derivatives to search for optimal values of the design variables \citet{dababneh2018application}. Although they possess fast convergence rates and can perform well for smooth problems with many design variables, one of the major disadvantages of gradient-based approaches is that global optimization is difficult to ensure when dealing with non-smooth problems with several local optima \citet{dababneh2018application}. In addition to this limitation, gradient-based optimizers do not apply to optimization problems where the functional form is not readily available, often referred to as black-box optimization (BBO). In such cases, we can observe the outputs of the function based on some given input parameters, but the form of the function is unavailable. On the other hand, derivative-free optimization methods that do not utilize gradients for optimization, as the name implies, have been developed. Among these are some evolutionary algorithms, loosely modeled after biological evolution, and swarm-based techniques, inspired by social behavior in biological swarms. Derivative-free optimizers possess advantages such as robustness to the occurrence of local optima, suitability for BBO problems, and ease of implementation (since analytical and numerical gradients are not required) \citet{houssein2021major}. In particular, swarm-based optimizers operate based on the principle of swarm intelligence (SI), which occurs in the collective intelligent behavior of decentralized and self-organized systems \citet{ab2015comprehensive}. Examples of swarm-based optimizers are Ant Colony Optimization (ACO) \citet{dorigo2006ant}, Artificial Bee Colony (ABC) \citet{karaboga2010artificial}, Cuckoo Search Algorithm (CSA) \citet{yang2010engineering}, Glowworm Swarm Optimization (GSO) \citet{krishnanand2009glowworm}, and Particle Swarm Optimization (PSO) \citet{eberhart1995particle, hu2003engineering, poli2007particle}. These algorithms typically operate by gradually updating each particle's position within the design space based on a rule that governs how it interacts with other swarm members. As opposed to each swarm member updating its position based on only its knowledge of the design space, the swarm members work together to find the global optimum, thus achieving better performance than attainable with a single member working alone.

Despite their success in optimizing non-convex multimodal functions, swarm-based optimizers typically suffer from slow convergence and require several function evaluations to find the global optima \citet{owoyele2021novel}. Also, in many engineering applications, evaluating the objective function can be expensive or time-consuming. For instance, searching a compositional parameter space for a new material requires following the complex set including experimental setup, sample preparation, material testing, etc. In many industries, computational fluid dynamics (and other types of simulation approaches) may be used to perform function evaluations, and depending on the numerical complexity, a single simulation may require several hours or days to complete on several processors. Therefore, for such applications with expensive function evaluations, design optimizers must be designed to work with limited swarm sizes, otherwise, the optimization process becomes impractical due to excessive computing runtimes. For problems where a small swarm size is necessarily imposed due to runtime-related considerations described above, swarm-based optimizers often suffer from premature convergence to local optima \citet{owoyele2021novel}. Another downside is that the performance of swarmed-based optimizers for various problems is often sensitive to the choice of model parameters or constants (e.g., maximum velocity, communication topologies, inertia weights, etc.). These constants are required to be selected by the user a priori and do not necessarily generalize adequately across problems \citet{shi1998parameter, trelea2003particle}. Thus, parameters need to be carefully tuned to maintain acceptable performance for different optimization problems \citet{houssein2021major}.

In this study, we introduce DeepHive, a new approach to developing swarm-based design optimizers, using a machine learning strategy known as reinforcement learning. Reinforcement learning is a branch of artificial intelligence where an agent learns an optimal policy by interacting with the environment for sequential decision-making problems \citet{sutton1998introduction}. Some previous studies have explored the possibility of enhancing design optimization using reinforcement learning \citet{xu2020reinforcement}. \citet{li2016learning} developed a framework for learning how to perform gradient-based optimization using reinforcement learning. \citet{xu2020reinforcement} presented a variant of PSO based on reinforcement learning methods with superior performance over various state-of-the-art variants of PSO.  In their approach, the particles act as independent agents, choosing the optimal communication topology using a Q-learning approach during each iteration. \citet{samma2016new} developed a reinforcement learning-based memetic particle swarm optimizer (RLMPSO), where each particle is subjected to five possible operations which are exploitation, convergence, high jump, low-jump, and local fine-tuning. These operations are executed based on the RL-generated actions. 
Firstly, our paper differs from the preceding studies in that the proposed approach discovers an optimization policy within a continuous action space. Secondly, as opposed to some previous studies that learn parameters within the framework of existing swarm-based approaches, DeepHive learns to optimize with minimal assumptions about the functional form of the particle update policy. Lastly, DeepHive displays acceptable generalization characteristics to unseen functions, as will be shown in section 4. The remainder of the paper is laid out as follows: section 2 gives a quick overview of reinforcement learning and its relationship to BBO, including a description of the proximal policy optimization (PPO) algorithm employed in this study. The proposed optimization technique is discussed in detail in section 3. A brief explanation of the benchmark objective functions tested, and the results obtained using DeepHive is presented and analyzed in section 4. The paper ends with some concluding remarks in section 6.

\section{Proposed approach}
\label{sec:headings}

\subsection{Reinforcement Learning}
Reinforcement learning (RL) involves an interaction between an agent and its environment, wherein the agent receives a reward (or punishment) based on the quality of its actions. The agent takes an action that alters the state of an environment, and a reward is given based on the quality or fitness of the action performed. Each agent tries to maximize its reward over several episodes of interaction with the environment, and by so doing, learns to select the best action sequence. Typically, the reinforcement learning problem is expressed as a Markov decision process (MDP). The state and action space in a continuous setting of a finite-horizon MDP is defined by the tuple $(S, A, S, p_0, p, c, Y)$, where $S$ represents the states, $A$ is the set of actions, and $p_0$ is the probability density function over initial states $p_0$ : $S \to \mathbb{R}^+$,
$p$ is the conditional probability density over successor states given the current state and action, $p$ : $S \times A \times S \to \mathbb{R}^+$, $c$ is the function that maps states to reward $c:S \to  \mathbb{R}$, and $y \in (0,1]$ is the discount factor \citet{li2016learning}. Specifically, we employ the PPO algorithm, which belongs to a class of RL methods called policy gradient methods \citet{schulman2017proximal}, to train the policy (denoted $\pi_\theta$) in this work. The policy gradient methods operate by directly optimizing the policy which is modeled by a parameterized function of $\theta$, i.e., $\pi_\theta (a|s)$ . The policy is then optimized using a gradient ascent algorithm, which seeks to maximize reward. In the policy gradient approach, the goal is to maximize the surrogate objective defined as
\begin{equation}
\mathcal{L}^{CPI}(\theta) = \hat{\mathbb{E}} \left[\frac{\pi_{\theta} (a|s)}{\pi_{\theta_{old}}(a|s)}\cdot \hat{A}\right] = \hat{\mathbb{E}} \left[r({\theta}) \hat{A} \right]
\end{equation}

where $r({\theta})=  \frac{\pi_{\theta} (a|s)}{\pi_{\theta_{old}}(a|s)}$ is the ratio of the current policy to an old policy (or baseline policy) and is called the probability ratio. The superscript “CPI” stands for “conservative policy iteration”, $\hat{\mathbb{E}}$ is the expectation or expected value, $\theta$ represents the weights or the policy network parameters, and $\hat{A}$ is the advantage which measures how good an action is in a given state. During training, excessive policy updates can sometimes occur, leading to the selection of a poor policy from which the agent is unable to recover. To reduce the occurrence of such actions, trust region optimization approaches, such as PPO, have been developed. PPO prevents excessive policy updates by applying a clipping function that limits the size of the updates as follows:
\begin{equation}
   \mathcal{L}^{CLIP}(\theta) = \hat{\mathbb{E}}\left[ \min\left(r(\theta) \hat{A}, \text{clip}\left(r(\theta), 1 - \epsilon, 1 + \epsilon\right) \hat{A}\right)\right] 
\end{equation}

The first term on the right-hand side of Eq. 2 is $\mathcal{L}^{CLIP}(\theta)$ as defined in Eq. 1. The second term, $(\text{clip}(r(\theta), 1 - \epsilon, 1 + \epsilon) \hat{A}$ is the modification which clips the probability ratio to the interval $[1 - \epsilon, 1 + \epsilon]$. The term $\epsilon$ is a hyperparameter that controls the maximum update size, chosen to be 0.2 in the original PPO paper \citet{schulman2017proximal}.

When the actor and critic function share the same neural network parameters, the critic value estimate is added as an error term to the objective function. Also, an entropy term is added for exploration, hence the final objective is given as:
\begin{equation}
c(\theta) = \hat{\mathbb{E}}[\mathcal{L}^{CLIP}(\theta) - c_{1}\mathcal{L}^{VF}(\theta) + c_{2}S[\pi_\theta](s) ]
\end{equation}

Where  $c_1$, $c_2$ are coefficients, and $S$ denotes the entropy bonus term for exploration, and $\mathcal{L}^{VF}(\theta)$  is a squared-error loss which is the squared difference between the state value and the cumulative reward. In previous studies, PPO has been applied to the optimization of a range of reinforcement learning problems, displaying good performance \citet{huang2020deep, schulman2017proximal, yu2021surprising}.

\subsection{RL-based Optimization Method}
Swarm-based optimizers can be thought of as a policy for updating the position of agents or particles within a design space. In general, swarm-based optimizers update the position of each swarm member based on the particles’ present and historical positions and the objective function values. In the study, we seek to develop an approach for discovering such policies using deep reinforcement learning.
First, we distinguish between two distinct optimization endeavors carried out in this study. The first involves the training of the reinforcement learning agents, where they \emph{learn} to find the peaks or lowest points of surfaces by several repeated attempts. During policy generation, the agents attempt to optimize a benchmark optimization function repeatedly as though they are playing a game, progressively getting better at finding the optimum. We refer to this hereafter as the \emph{policy generation} phase, which is obtained using the PPO algorithm as described in section 2.1. The policy generation phase yields a policy, $\pi_\theta$, that can be used to perform global optimization of practical problems. The policy, $\pi_\theta$, is computed using a deep neural network that takes information about the state of the swarm and outputs an action. 
The second phase involves \emph{global optimization} or \emph{policy deployment}, where the policy discovered in the first phase is applied to the optimization of an objective function. Here, agents are randomly initialized over the entire design space at the initial design iteration. Afterward, the location of the agents is progressively adjusted using the generated policy, until they cooperatively find the design optimum. The update vector used to adjust the position of the particles is computed based on the policy, $\pi_\theta$. The update function $\pi_\theta$ in this study, like swarm-based optimizers, is a function of the current and prior states of agents within the swarm parameterized by $\theta$  which are the weights of a neural network. The definitions of elements of the MDP are as follows. The state is made up of the agents' present and previous locations within the design space, as well as the objective function values at those locations. The action is a vector that specifies the distance between the agent’s current location and its new location based on the application of policy, $\pi_\theta$. The reward is a numeric value assigned based on how good the current policy is at performing global optimization. These two elements, namely policy generation and deployment are discussed in more detail below.

\begin{figure}[ht]
\centering
\includegraphics[scale=0.7]{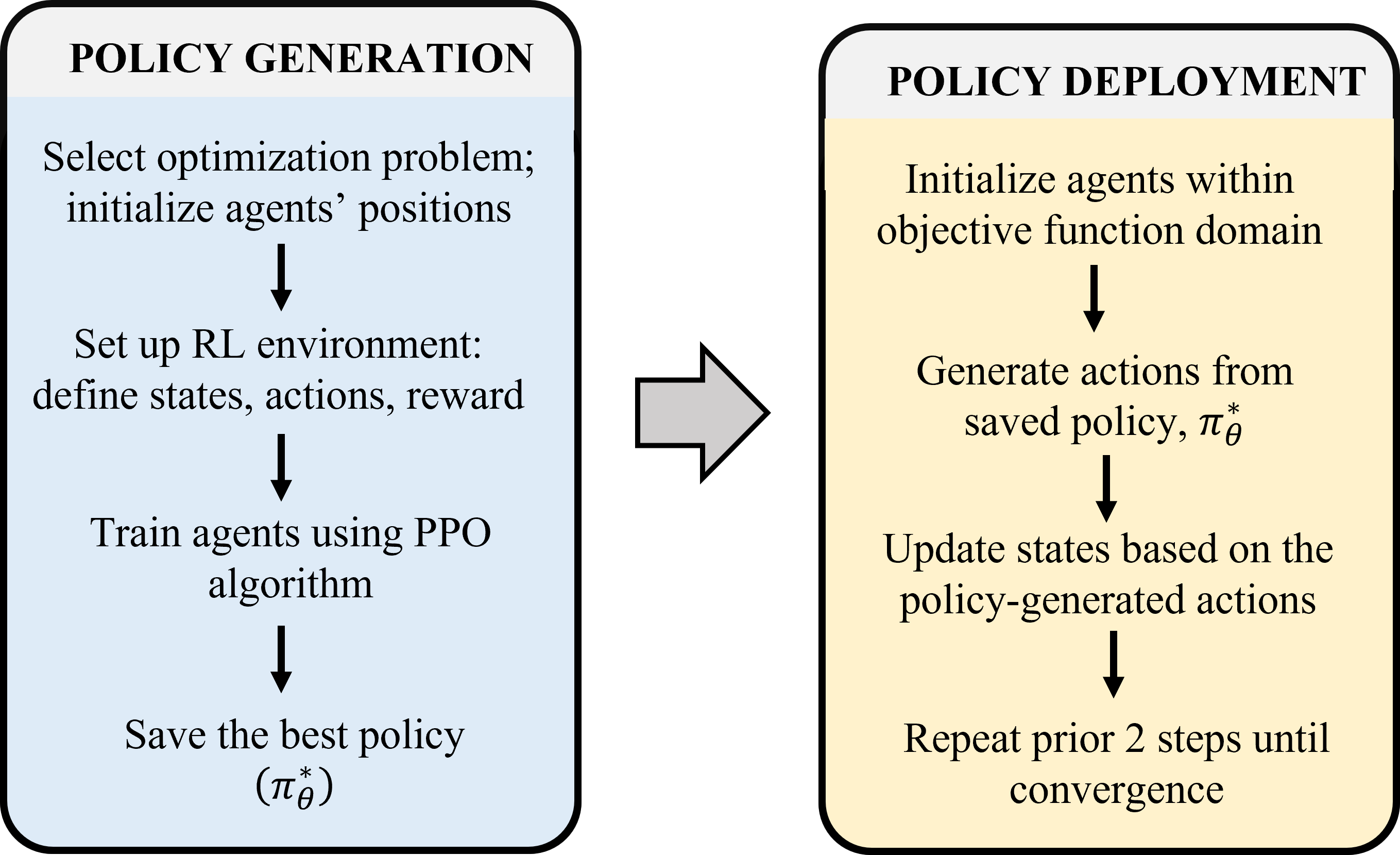}
\caption{Illustration of policy generation and deployment phases of DeepHive.}
\end{figure}

\subsubsection{Policy Generation}
As with every reinforcement learning problem, the environment is the agent’s world. An agent cannot influence the rules of the environment by its actions. Nevertheless, it can interact with the environment by performing actions within the environment and receiving feedback in the form of a reward. In this work, the environment is customized from the Open-AI gym library \citet{brockman2016openai}. The problem is framed as an episodic task, where each episode consists of 25 updates to the state of the swarm (i.e., design iterations), after which the optimization attempt is terminated, regardless of whether the swarm converges or not. At the beginning of each episode, the agents take random positions within the domain of interest. These positions and objective function values are scaled to fall within a [0, 1] interval, which enables the resulting policy to apply to problems where the order of magnitude of the domain bounds and objective function are different. However, for BBO, the minimum and maximum values of the objective function are not known \emph{a priori}. Therefore, at a given iteration, the range chosen for normalization is based on the currently known global best and global worst objective function values. 
We denote the position vector for the $i^{th}$ agent in a swarm of $N$ agents as  $x_i=[x_{(i,1)},x_{(i,2)},…,x_{(i,D)} ]$, where $D$ is the number of dimensions (or independent variables) in the objective function, $f$. The elements of $\textbf{x}_i$, denoted $x_{(i,j)}$, represent the position of agent $i$ in the $j^{th}$ dimension. Finally, $\bar{\textbf{x}}_i$ is a vector that denotes the historical best position of agent $i$, i.e., the location of the maximum objective value that agent $i$ has reached in since the swarm was initialized. The observation vector for agent $i$ in the $j^{th}$ dimension is given by:

\begin{equation}
   \boldsymbol\zeta_{i, j} = [(x_{i,j} - \hat{x}_{i,j}), (f(\textbf{x}_{i}) - f(\bar{\textbf{x}}_i)),(x_{i,j} - x_{n, j}),(f(\textbf{x}_{i}) - f(\textbf{x}_{n})) 
\end{equation}

At every training epoch, the observation, $\boldsymbol\zeta$, is fed to the actor neural network, which outputs a mean action. The agents’ actions are then sampled from a distribution created with the mean action and a set standard deviation. The policy generation has two phases. During the first stage, the environment is explored more aggressively, so that the agents are exposed to a wide range of states within the environment. During this stage, the position of agent $i$ in the $j^{th}$ dimension, arbitrary agent, $x_{i,j}$, is updated using a normal distribution with the mean obtained from the PPO policy and a fixed standard deviation of 0.2.
\begin{equation}
    \Delta{x}_{i,j} = \mathcal{N}(\pi_{\theta}(\boldsymbol\zeta_{i,j}), 0.2)
\end{equation}

Eq. 5 is used to update the position vector during the first 2500 iterations. Afterward, to update the policy at a given iteration, we compute the action of an agent by sampling from a normal distribution with the mean as the output of the policy network and standard deviation which is a function of the particle’s distance from the global best particle’s distance: 
\begin{equation}
    \Delta{x}_{i,j} = \mathcal{N}(\pi_{\theta}(\boldsymbol\zeta_{i,j}), \psi(x_{i,j} - x_{g,j}))
\end{equation}
Where $g$ is the index of the globally best agent $\psi:\mathbb{R} \to \mathbb{R}$ is a linear function given by: 
\begin{equation}
    \psi(x_{i,j} - x_{g,j}) = 0.002 + 0.18(x_{i,j} - x_{g,j})
\end{equation}

In Eq. 7, the standard deviation for the update of agent $i$ a linear function of its distance from the agent that has the best location globally. Thus, the stochastic component of a particle’s movement is large when far away from the global best to encourage exploration, but its update becomes more deterministic as the design optimum is approached to promote better exploitation. At each time-step, the RL agents take actions which are steps within the domain of the objective function. The reward function is designed such that actions that lead to subsequently poor objective value are penalized, while those that lead to adequate and fast convergence are rewarded. The reward function used in this paper is:
\begin{equation}
    \mathcal{R} = \begin{cases} 
10(f(x^{(k)}) - f(x^{(k-1)})) & \text{if } f(x^{(k)}) < \kappa \\
10[1 + (f(x^{(k)}) - f(x^{(k-1)}))] & \text{otherwise} \\
\end{cases}
\end{equation}

Eq. 8 assumes a maximization problem, where the superscript, $k$, denotes the iteration number. $k$ is defined as the reward transition number and should be chosen to be close to the best-known optimum (0.9 in this paper). In other words, when the objective function value of an agent is far from the globally best objective function value, the reward is proportional to the difference in objective function value between the current and previous locations. The reward is positive if the agent moves to a better location, and negative if it moves to a worse location. Once an agent crosses the reward transition number, it gets an additional reward of 10 at each training epoch. This serves two purposes. First, this encourages the RL agents to quickly move to regions close to the design optimum and remain there, since more rewards are reaped in such regions. Secondly, rewards become more difficult to accumulate close to the global optimum, since there is limited room for improvement to the current objective function value. Thus, the additional reward in regions close to the global best helps establish a baseline reward, such that particles in this region are adequately rewarded. 
By learning to maximize reward, the agents learn to take actions that lead to high rewards, namely, actions that find regions close to the global optimum on the surface. In this study, all the agents used for policy generation share the same policy for choosing their actions within the design space. Furthermore, each dimension within the design space is updated separately. Therefore, the number of dimensions and the agents can be readily varied during model deployment to solve a global optimization problem. Policy generation is performed using the cosine mixture function (section 3.1) and 7 agents. The policy is modeled with a neural network consisting of 2 hidden layers, each containing 64 neurons and activated using the hyperbolic tangent function. The training was performed for 250,000 iterations (which is equivalent to 10,000 episodes).
\subsubsection{Policy Deployment}
The reinforcement learning approach for policy generation is summarized in algorithm 1. Once the training process described in section 2.2.1 is complete, the best policy $(\pi_{\theta}^{*})$ is saved, and afterward, can be deployed for global optimization of functions. The steps involved in the policy deployment stage are summarized in Algorithm 1. As in the policy generation phase, the positions of the particles are progressively updated. In contrast to policy generation, the policy is frozen during this process. Furthermore, the best particle at each design iteration is kept frozen. This provides an anchor for the entire swarm and prevents all the particles from drifting off from promising regions of the design space.

\begin{algorithm}
    \caption{Reinforcement Learning-based Optimization}
    \begin{algorithmic}
        \Require number of dimensions $D$; number of agents, $N$; maximum number of iterations, $K$
        \Require objective function, $f$ ; exploration function, $\psi$
        \Statex Define: agent index, $i$; dimension index, $j$; iteration number, $k$; index of globally best agent, $g$
        \State Initialize $\textbf{x}_1^{(0)}, \textbf{x}_2^{(0)}, ..., \textbf{x}_N^{(0)}  $
        \Comment {Randomly initialize position vector of agents 1 to $N$, where $\textbf{x}_i = [x_{i,1}, x_{i,2},..., x_{i,D}]$}
        \State $\bar{\textbf{x}}_i = \textbf{x}_i$ for $i = 1,2, ..., N$ \Comment{Initialize all agents best position}
        \For{$k = 1,2,...,K$} \Comment{loop through iteration number}
        \State $g = argmax\{f(x_1^{(k)}), f(x_2^{(k)}), f(x_N^{(k)}) \}$ \Comment{get the index of the globally best agent}
        \For{$i=1,2,...,N$ $i\not=g$}\Comment{loop through the number of agents excluding the best agent}
        \State $n = rand(1,2,...,N)\not=i$\Comment{get randomly selected neighbor}
        \For{$j=1,2,...,D$}\Comment{loop through dimension number}
        \State $\boldsymbol\zeta_{i,j} = [(x_{i,j}^{(k)} - \bar{x}_{i,j}), (f(\textbf{x}_{i}^{(k)}) - f(\bar{\textbf{x}}_{i})),(x_{i,j}^{(k)} - x_{n,j}^{(k)}), (f(\textbf{x}_{i}^{(k)}) - f(\textbf{x}_{i}^{(k)}))]$
        \State $\Delta x_{i,j}^{(k)} \gets \mathcal{N}(\pi_\theta(obs_{i,j}), \psi(x_{i,j}^{(k)}, x_{g,j}^{(k)})$\Comment{Evaluate change in agent's position}
        \State $x_{i,j}^{(k+1)} \gets x_{i,j}^{(k)} + \Delta x_{i,j}^{(k)}$\Comment{Update the position of the agents}
        \EndFor
        \If{$f(\textbf{x}_{i}^{(k+1)}) > f(\bar{\textbf{x}}_{i})$}\Comment{Update agent i best position}
            \State $\bar{\textbf{x}}_{i} = \textbf{x}_{i}^{(k+1)}$
        \EndIf
    \EndFor 
\EndFor 

\end{algorithmic}
\end{algorithm}

\section{Results and Discussion}
\label{sec:others}

\subsection{Performance of DeepHive on the cosine mixture function}
\begin{figure}[H]
\centering
\includegraphics[scale=0.65]{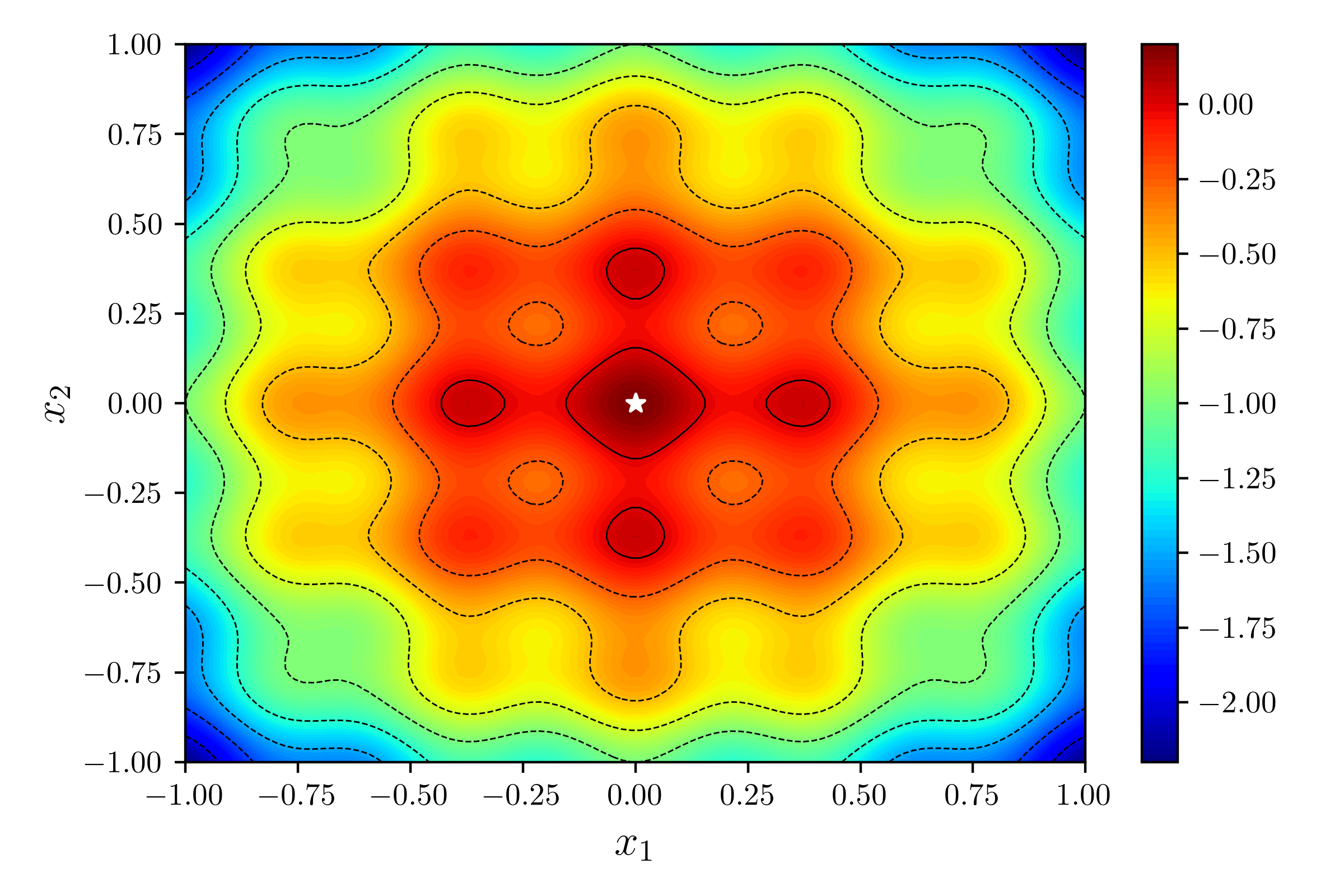}
\caption{Contour plot of cosine mixture function. The white star in the figure represents the location of the global maximum.}
\end{figure}

In this section, we compare the performance of DeepHive with PSO \citet{eberhart1995particle}, differential evolution (DE) \citet{storn1997differential}, from the python SciPy library \citet{virtanen2020scipy} and GENetic Optimization Using Derivatives (GENOUD) \citet{mebane1997genetic} algorithm. PSO is a swarm-based optimizer, while DE and GENOUD are both based on genetic algorithms. First, we evaluate the performance of DeepHive using identical scenarios to training. In other words, we deploy the policy by testing it with the same objective function, number of particles, and number of dimensions, and therefore, these results depict an ideal performance. The training was performed using 10 particles on the 2-dimensional (2D; i.e., D = 2) cosine mixture function \citet{ali2005numerical} based on the procedure described in section 2.2.1. The cosine mixture function is shown in Fig. 2. It is a multi-modal function consisting of 25 local maxima, with one of these being the global maximum. The objective function value, f, is defined as a function of design parameters, x, as 
\begin{equation}
f(\mathbf{x}) = 0.1\sum_{j=1}^{D}\cos(5 \pi x_j) - \sum_{j=1}^{D}x_j^{2}
\end{equation}

subject to $-1<=x_j<=1$ for $j = 1,2,...,D$. The global maximum is $f_{max}(x^*) = 0.1D$ located at $x^* = 0$.

\begin{figure}[H]
\centering
\includegraphics[scale=0.65]{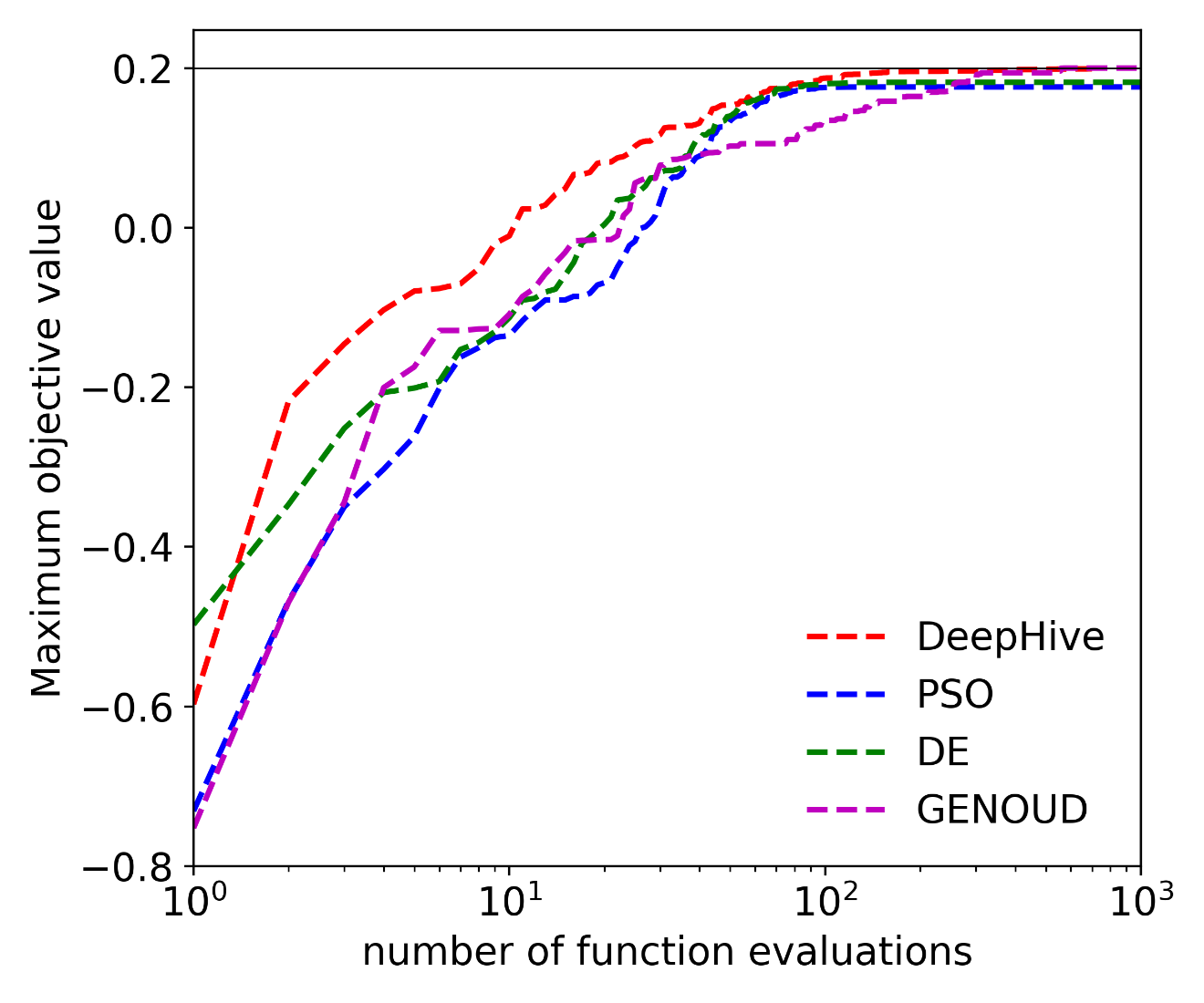}
\caption{cosine mixture function contour (a), RL-optimizer compared with other optimizers (b), performance of RL-optimizer for various numbers of agents (c), Generalization test to higher dimensions (d) }
\end{figure}

\begin{table}[ht]
    \centering
    \caption{Detailed comparison of DeepHive’s with other optimizers for the cosine mixture function}
    \begin{tabular}{|c|c|c|c|c|}
        \hline
        Iteration Number, k & DE & PSO & GENOUD & DeepHive \\
        \hline
        0 & -0.4975±0.3836 & -0.7312±0.4345 & -0.7525±0.4417 & -0.4045±0.4534 \\
        100 & 0.1803±0.0489 & 0.1757±0.054 & 0.1345±0.0729 & 0.1833±0.0303 \\
        200 & 0.1822±0.048 & 0.1764±0.0542 & 0.1645±0.0631 & 0.1925±0.0147 \\
        300 & 0.1823±0.048 & 0.1764±0.0542 & 0.1878±0.04 & 0.1958±0.0069 \\
        400 & 0.1823±0.048 & 0.1764±0.0542 & 0.1941±0.029 & 0.1975±0.0034 \\
        500 & 0.1823±0.048 & 0.1764±0.0542 & 0.1941±0.029 & 0.1985±0.0013 \\
        600 & 0.1823±0.048 & 0.1764±0.0542 & 0.2±0.0 & 0.1988±0.001 \\
        700 & 0.1823±0.048 & 0.1764±0.0542 & 0.2±0.0 & 0.199±0.0008 \\
        800 & 0.1823±0.048 & 0.1764±0.0542 & 0.2±0.0 & 0.199±0.0008 \\
        900 & 0.1823±0.048 & 0.1764±0.0542 & 0.2±0.0 & 0.1992±0.0008 \\
        1000 & 0.1823±0.048 & 0.1764±0.0542 & 0.2±0.0 & 0.1999±0.0002 \\
        \hline
    \end{tabular}
    
    \label{tab:optimization_results}
\end{table}

Figure 3 shows a comparison of the performance of DeepHive compared to PSO, DE, and GENOUD. In the figure, the best objective function value is plotted against the number of function evaluations. In Fig. 3 (and all the remaining plots in the results section), the plots show the mean performance across 25 trials. In other words, the process of initialization and optimization described in Algorithm 1 was repeated 25 times, and the performance was averaged, to provide a statistically valid projection of real-world performance. From the figure, it can be seen that on average, DeepHive requires a lower number of function evaluations to reach the vicinity of the optimum design. PSO and DE, in particular, fail to reach the optimum, only achieving a maximum fitness value of $0.1764 \pm 0.0542$ and $0.1823 \pm 0.048$ respectively, even after 100 iterations. While GENOUD converged to the global optimum, the maximum function value obtained using DeepHive rises faster compared to GENOUD. This is particularly important for expensive function evaluations in applications where a high degree of accuracy in locating the design optimum is less critical than function evaluation efficiency. For instance, setting a threshold of $0.19$, we see that it takes DeepHive 130 function evaluations to reach this function value, while GENOUD takes 350 function evaluations, on average. Table I shows numeric values of the averaged maximum fitness obtained by all the optimizers with the standard deviations included (mean ± standard deviation). For all entries, the standard deviation is rounded to four decimal places. We see that DeepHive consistently outperforms all the other optimizers, except after 500 function evaluations, where the local search component of GENOUD leads to minor improvements in the objective function compared to DeepHive. Due to the absence of a dedicated local search component, the maximum objective function for DeepHive peaks at $0.1999$, with a difference of $0.0001$ from the global optimum. 

\subsection{Effect of Number of Particles}

\begin{figure}[H]
\centering
\includegraphics[scale=.65]{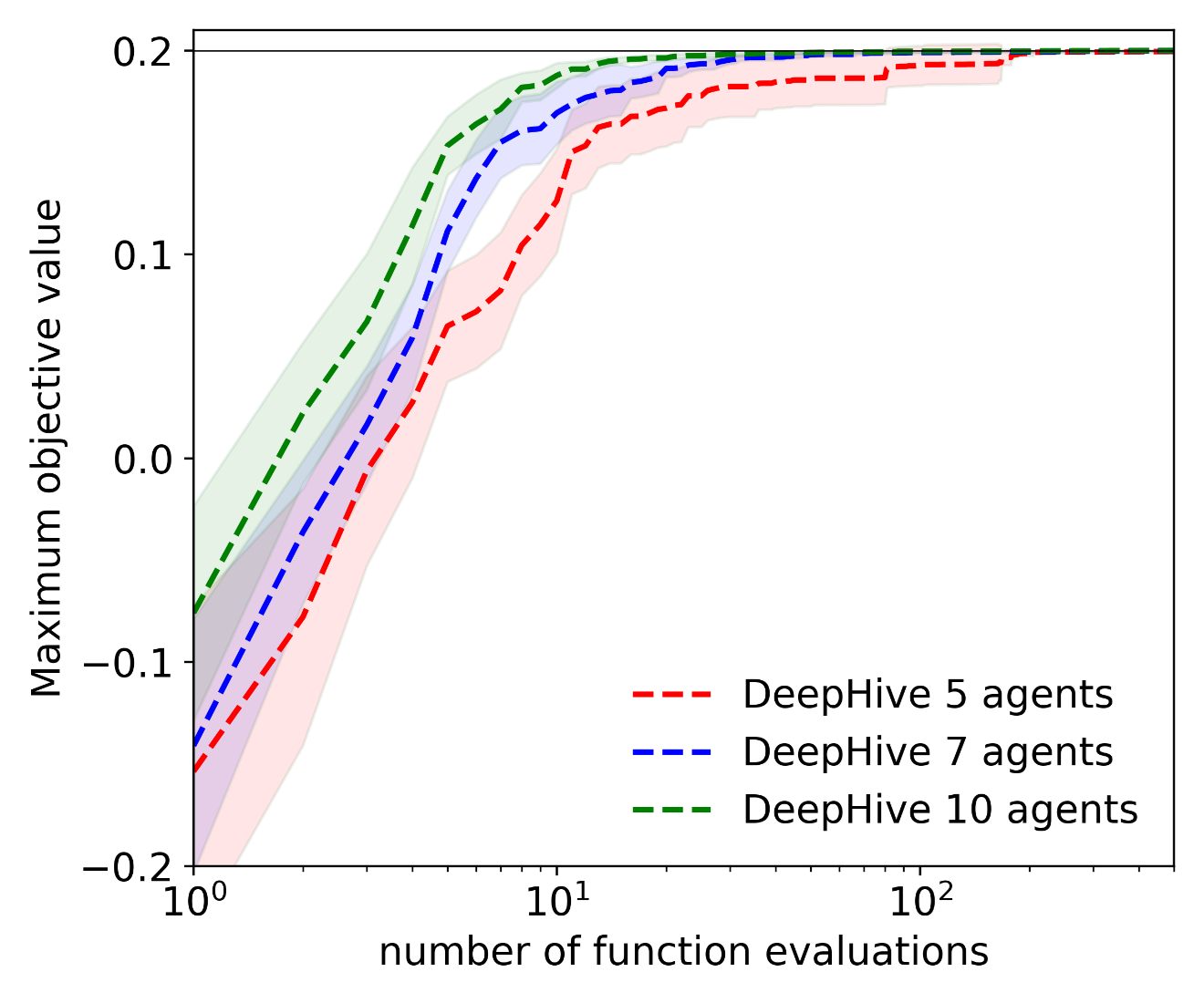}
\caption{The performance of DeepHive for various numbers of agents. The horizontal black line represents the maximum achievable objective value.}
\end{figure}

Next, we discuss the effect of the number of agents used. As mentioned in section 2.2.1, all agents used during the policy generation phase share the same policy for updating their positions. Therefore, the number of agents can be scaled up or down after training, depending on the application and cost of the function evaluations. Here, we compare the use of 5, 7, and 10 agents, and as done in 3.1, Fig. 4 shows the maximum objective value as a function of the number of function evaluations. Figure 4 depicts a direct comparison of the performance of 5, 7, and 10 agents, demonstrating that the number of agents involved in the search is directly related to the speed of convergence. Here, the solid dashed lines represent the mean performance over 25 trials, while the shaded regions represent the standard deviation. As desired, DeepHive’s performance shows positive scaling with the number of particles. First, we note that DeepHive successfully finds the design optimum with 5 agents, despite training being conducted with 7 agents. Secondly, the performance improves as the number of agents increases. Thus, the policy generated takes advantage of additional information provided by 10 agents, though it uses a policy generated by training 7 agents. In this case, 10, 7, and 5 agents reached a function value of 0.19 after 100, 130, and 400 function evaluations, respectively.

\subsection{Effect of Dimensionality}
\begin{figure}[H]
\centering
\setlength{\textfloatsep}{10pt plus 1.0pt minus 2.0pt}
\includegraphics[scale=0.65]{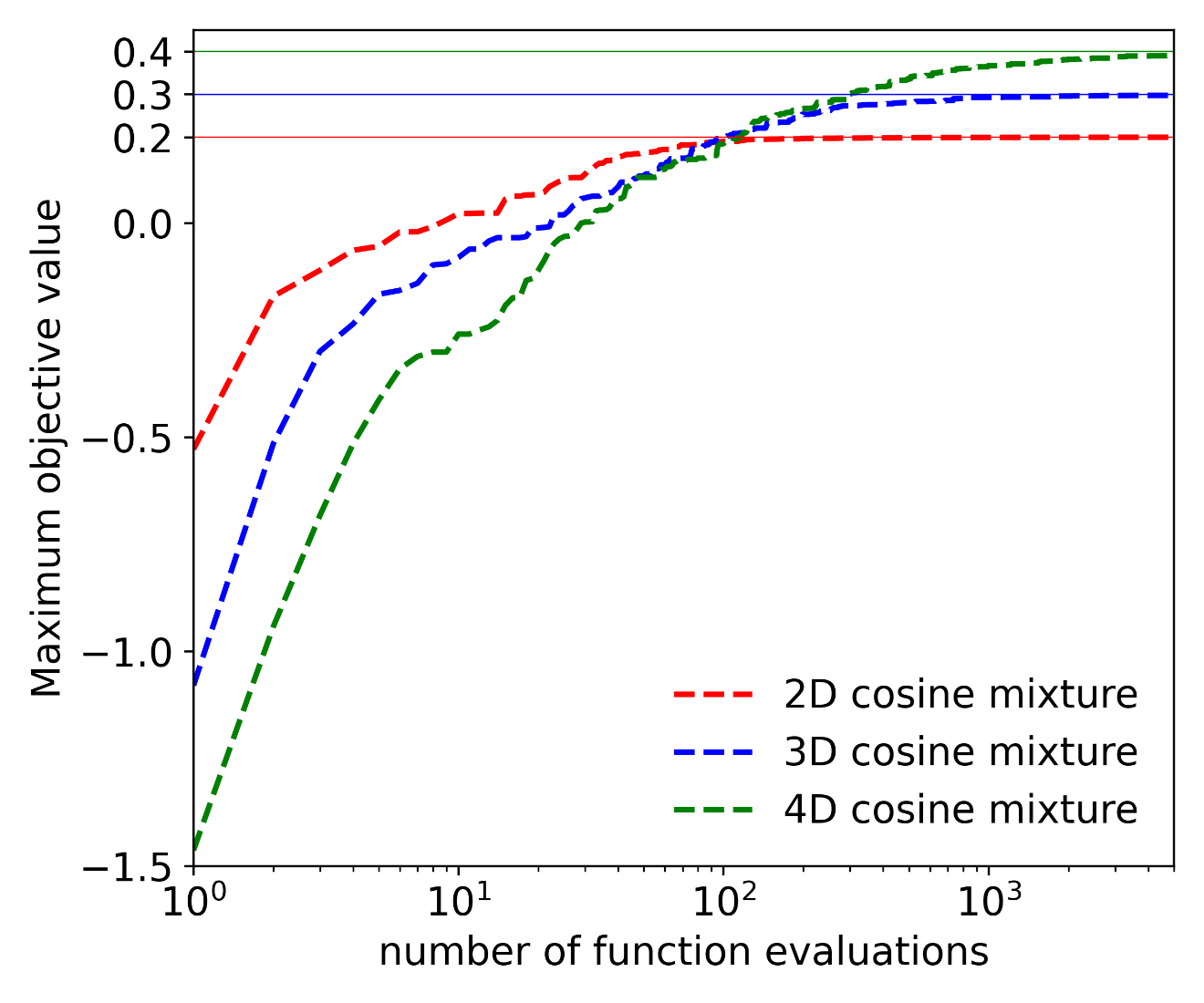}
\caption{The performance of DeepHive for various numbers of agents. The horizontal black line represents the maximum achievable objective value.}
\end{figure}

Since the updates are applied to each dimension within the design space separately, DeepHive can be applied to problems with different dimensions than it was trained with. Next, we investigate the capability of the trained policy to generalize to higher dimensions. We do this by testing the policy generated using a 2D cosine mixture on 3D and 4D versions of the same function. This comparison is shown in Fig. 5, where 10 agents have been used for policy deployment in all cases. As expected, the number of iterations needed to reach the region around the global maximum increases as the dimensionality increases. This is in part due to the number of local maxima that scales exponentially with the dimensionality of the problem. Specifically, the number of local maxima for the cosine mixture problem is $5^D$, where D is the number of dimensions. As a result, the 4D cosine mixture problem has a 625 local maximum, as opposed to 125 for the 3D case, and 25 for the 2D case. Another reason for the decreased performance of the optimizer relates to the exponential growth of the design space was the number of dimensions increases. Nonetheless, we see that DeepHive can reach the vicinity of the design optimum using 10 agents, albeit taking longer for the higher dimension (as expected). The maximum achievable function value for the 2D, 3D, and 4D cases are 0.2, 0.3, and 0.4, respectively. For the 4D case, DeepHive reaches a function value of 0.37 after 1500 function evaluations, increasing to 0.3906 after 5000 iterations. For the 3D case, it reaches 0.29 after 800 iterations, increasing to 0.2978 after 4500 iterations. 

\section{Generalization to other functions}
We developed a multimodal test global optimization function based on the cosine function to evaluate DeepHive’s generalization abilities. This function has 15 local optimal solutions and 1 global optimal solution as shown in Fig. 6a. Additionally, we experimented with three additional global optimization test functions. It should be noted that all the functions tested in this subsection are all unseen by the agents since training (or policy generation) was carried out using the cosine mixture function only. Contour plots of these test functions are shown in Figs. 6b-d, while their mathematical descriptions are presented below.

\begin{enumerate}
    \item Function 1: Multi-modal cosine function with 15 local maxima and 1 global maximum 
    \begin{equation}
    f(\textbf{x}) = \sum_{j=1}^{D}cos(x_j - 2) + \sum_j^{D}cos(2x_j-4) + \sum_j^{D}cos(4x_j - 8) 
    \end{equation}
    subject to $-10<=x_j<=10$ for $j = 1,2,...,D$. The global maximum is $f_{max}(x^*) = 6$ located at $x^* = (1, 3)$.
    \item Function II \citet{gbhat}: Non-convex function with two local maxima (including one global maximum)
    \begin{equation}
        f(\textbf{x}) = (1 - \frac{x_1}{2} + x_1^{5} + x_2^{3}) \times  e^{-x_1^{2} - x_2^{2}}
    \end{equation}
    subject to $-10<=x_j<=10$ for $j = 1,2,...,D$. The global maximum is $f_{max}(x^*) = 1.058$ located at $x^* = (-0.225, 0)$.
    \item Matyas function \citet{hedar2007global}:
    \begin{equation}
        f(\textbf{x}) = 0.26(x_1^{2} + x_2^{2}) - 0.48x_1 x_2
    \end{equation}
    subject to $-10<=x_j<=10$ for $j = 1,2,...,D$. The global maximum is $f_{max}(x^*) = 0$ located at $x^* = (0, 0)$.
    \item Six hump camel function \citet{molga2005test}:
    \begin{equation}
         f(\textbf{x}) = (4 -  2.1x_1^{2} + \frac{x_1^{4}}{3}) + (-4+4x_2^{2})x_2^{2}
    \end{equation}
    subject to $-3<=x_j<=3$ for $j = 1,2,...,D$. The global maximum is $f_{max}(x^*) = -1.0316$ located at $x^* = (0.0898, -0.7126) or (-0.0898, 0.7126)$.
    
\end{enumerate}
\begin{figure}[ht]
\centering
\includegraphics[scale=0.55]{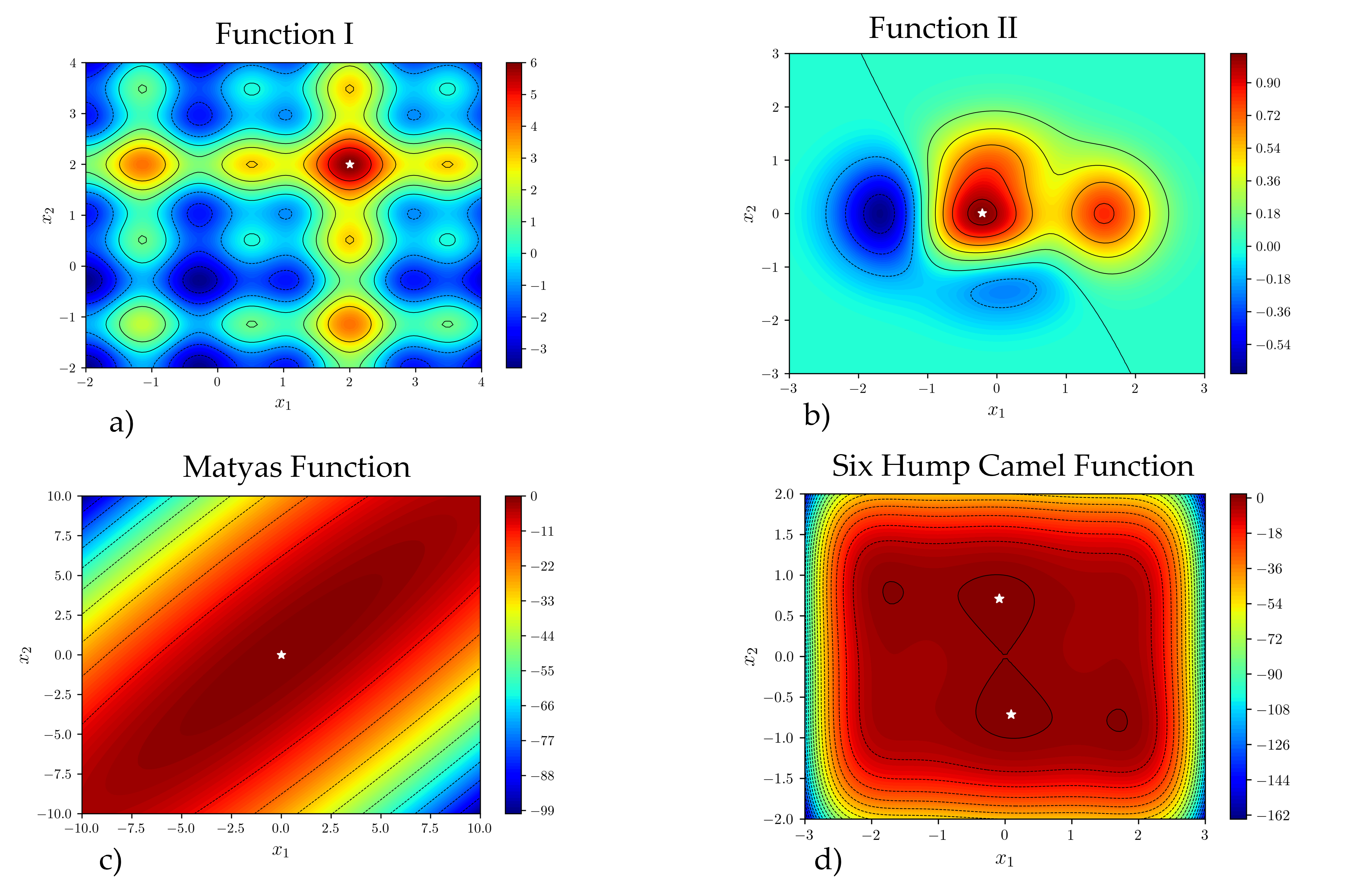}
\caption{The performance of DeepHive for various numbers of agents. The horizontal black line represents the maximum achievable objective value.}
\end{figure}
A comparison of DeepHive with other optimizers is shown in Figs. 7a-d, where the function value against the number of function evaluations is plotted. Figure 7a depicts the results of function I. At 250 function evaluations, DeepHive outperformed other optimizers reaching a fitness value of $5.9151 \pm 0.1773$, while PSO, DE, and GENOUD had a fitness value of $5.6999\pm0.7141$, $5.7596 \pm 0.6498$, $5.84 \pm 0.5426$ respectively. After 800 function evaluations, only DeepHive and GENOUD appear to reach the global maximum’s vicinity consistently, with the other optimizers (PSO and DE) sometimes getting stuck in local maxima. The result of function II, which has a non-convex surface with one global maximum and a local maximum just beside the global maximum, is shown in Fig. 7b. In this test, DeepHive also has a better performance compared to PSO and DE but was narrowly outperformed by GENOUD. At 300 function evaluations, DeepHive had a fitness value of $1.0503 \pm 0.0213$, with PSO, DE, and GENOUD having $ 1.0207 \pm 0.0832$, $1.0116 \pm 0.0908$, $1.0571 \pm 0.0$, respectively. At 1000 function evaluation, only DeepHive and GENOUD had reached the vicinity of the global maximum with a fitness value of $1.0566 \pm 0.0008$ and $1.0571 \pm 0.0$, respectively.

\begin{figure}[H]
\centering
\includegraphics[scale=.65]{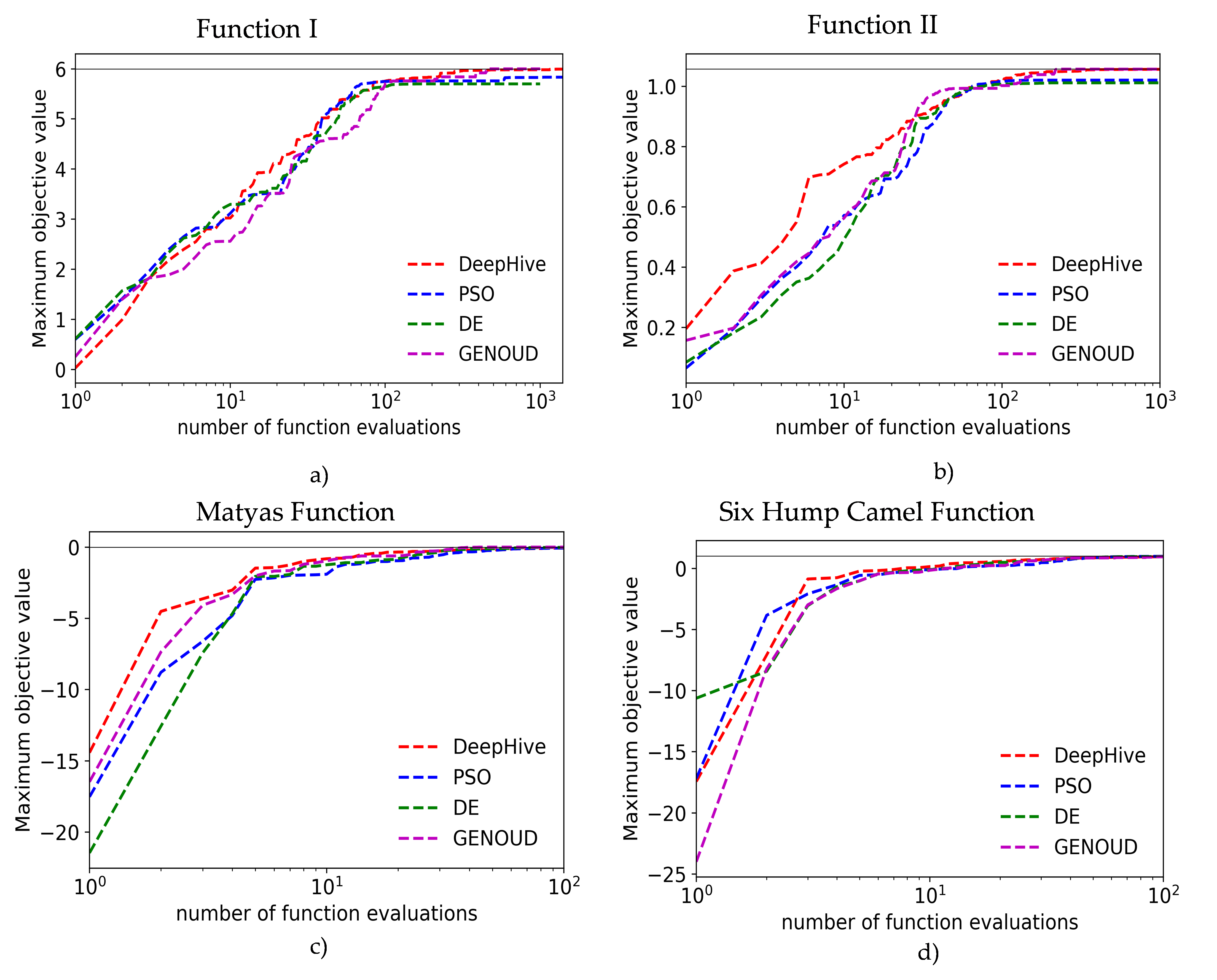}
\caption{The performance of DeepHive for various numbers of agents. The horizontal black line represents the maximum achievable objective value.}
\end{figure}

The result of the Matyas function, which is a smooth function with the global maximum $(f_{max}=0)$ at the center of the search space is presented in Fig. 7c. After 50 function evaluations, DeepHive reached an appreciable fitness value of $-0.0282 \pm 0.0315$ while PSO and DE have fitness values of $-0.224 \pm 0.3717$ and $-0.1121 \pm 0.1801$, respectively. GENOUD had a better performance as it converged to the global maximum within 50 function evaluations as in the cosine mixture function (section 3.1), DeepHive rapidly reaches the vicinity of the global maximum, before GENOUD becomes the highest performing optimizer due to its more intensive local search. For example, after 30 function evaluations, DeepHive attains a maximum function value of $-0.0738 \pm 0.0759$, while GENOUD only reaches $-0.2333 \pm 0.2752$. In many practical engineering applications, refinements close to the global optimum may not be as critical as function evaluation efficiency, since the uncertainty in measurements and simulations may be much higher than the refinements performed by GENOUD. In contrast to the Matyas function, the six-hump camel function is a minimization problem that features steep contours around its bounds and contains two global minima. For this function (Fig. 7b), after 500 function evaluations, the maximum fitness of DeepHive is $1.0286 \pm 0.0026$, compared with $0.9823 \pm 0.1798$, $1.0281 \pm 0.0161$, and $1.0316 \pm 0.0$ for PSO, DE, and GE, respectively. DeepHive converged at $1.0302 \pm 0.0009$ while PSO, DE, and GENOUD converged at $0.9823 \pm 0.1798$, $1.0314 \pm 0.0009$, $1.0316 \pm 0.0$ respectively. 
The Appendix section contains tables comparing the DeepHive algorithm's performance to that of other optimization algorithms for these benchmark functions.

\section{Conclusion}
In the paper, we presented a reinforcement learning-based approach for generating optimization policies for global optimization. In this approach, multiple agents work cooperatively to perform global optimization of a function. The proposed approach was compared with a particle swarm optimizer (PSO), direct evolutionary algorithm (DE), and GENetic algorithm using derivatives (GENOUD). The effect of changing the number of agents and dimensions was investigated, as well as the potential of generalization to other functions. The results showed good scaling performance with the number of agents, and that global optimization was still achievable even with higher dimensions. Most importantly, the proposed algorithm was able to locate the vicinity of the global optimum, even for functions it was not trained with, demonstrating generalization within the limited test functions used in this paper. Overall, DeepHive bears promise for use in domain-specific optimizers, where agents can be trained using a function generated using a low-fidelity, cost-effective model, then deployed on the full-order, expensive function. Future work will explore the effects of using multiple functions for training, the incorporation of local search components, and further validation studies on practical engineering problems.

\bibliographystyle{unsrtnat}
\bibliography{references}  

\begin{thebibliography}{29}
\providecommand{\natexlab}[1]{#1}
\providecommand{\url}[1]{\texttt{#1}}
\expandafter\ifx\csname urlstyle\endcsname\relax
  \providecommand{\doi}[1]{doi: #1}\else
  \providecommand{\doi}{doi: \begingroup \urlstyle{rm}\Url}\fi

\bibitem[Liao(2010)]{liao2010two}
T~Warren Liao.
\newblock Two hybrid differential evolution algorithms for engineering design
  optimization.
\newblock \emph{Applied Soft Computing}, 10\penalty0 (4):\penalty0 1188--1199,
  2010.

\bibitem[Dababneh et~al.(2018)Dababneh, Kipouros, and
  Whidborne]{dababneh2018application}
Odeh Dababneh, Timoleon Kipouros, and James~F Whidborne.
\newblock Application of an efficient gradient-based optimization strategy for
  aircraft wing structures.
\newblock \emph{Aerospace}, 5\penalty0 (1):\penalty0 3, 2018.

\bibitem[Houssein et~al.(2021)Houssein, Gad, Hussain, and
  Suganthan]{houssein2021major}
Essam~H Houssein, Ahmed~G Gad, Kashif Hussain, and Ponnuthurai~Nagaratnam
  Suganthan.
\newblock Major advances in particle swarm optimization: theory, analysis, and
  application.
\newblock \emph{Swarm and Evolutionary Computation}, 63:\penalty0 100868, 2021.

\bibitem[Ab~Wahab et~al.(2015)Ab~Wahab, Nefti-Meziani, and
  Atyabi]{ab2015comprehensive}
Mohd~Nadhir Ab~Wahab, Samia Nefti-Meziani, and Adham Atyabi.
\newblock A comprehensive review of swarm optimization algorithms.
\newblock \emph{PloS one}, 10\penalty0 (5):\penalty0 e0122827, 2015.

\bibitem[Dorigo et~al.(2006)Dorigo, Birattari, and Stutzle]{dorigo2006ant}
Marco Dorigo, Mauro Birattari, and Thomas Stutzle.
\newblock Ant colony optimization.
\newblock \emph{IEEE computational intelligence magazine}, 1\penalty0
  (4):\penalty0 28--39, 2006.

\bibitem[Karaboga(2010)]{karaboga2010artificial}
Dervis Karaboga.
\newblock Artificial bee colony algorithm.
\newblock \emph{scholarpedia}, 5\penalty0 (3):\penalty0 6915, 2010.

\bibitem[Yang and Deb(2010)]{yang2010engineering}
Xin-She Yang and Suash Deb.
\newblock Engineering optimisation by cuckoo search.
\newblock \emph{International Journal of Mathematical Modelling and Numerical
  Optimisation}, 1\penalty0 (4):\penalty0 330--343, 2010.

\bibitem[Krishnanand and Ghose(2009)]{krishnanand2009glowworm}
KN~Krishnanand and Debasish Ghose.
\newblock Glowworm swarm optimisation: a new method for optimising multi-modal
  functions.
\newblock \emph{International Journal of Computational Intelligence Studies},
  1\penalty0 (1):\penalty0 93--119, 2009.

\bibitem[Eberhart and Kennedy(1995)]{eberhart1995particle}
Russell Eberhart and James Kennedy.
\newblock Particle swarm optimization.
\newblock In \emph{Proceedings of the IEEE international conference on neural
  networks}, volume~4, pages 1942--1948. Citeseer, 1995.

\bibitem[Hu et~al.(2003)Hu, Eberhart, and Shi]{hu2003engineering}
Xiaohui Hu, Russell~C Eberhart, and Yuhui Shi.
\newblock Engineering optimization with particle swarm.
\newblock In \emph{Proceedings of the 2003 IEEE Swarm Intelligence Symposium.
  SIS'03 (Cat. No. 03EX706)}, pages 53--57. IEEE, 2003.

\bibitem[Poli et~al.(2007)Poli, Kennedy, and Blackwell]{poli2007particle}
Riccardo Poli, James Kennedy, and Tim Blackwell.
\newblock Particle swarm optimization.
\newblock \emph{Swarm intelligence}, 1\penalty0 (1):\penalty0 33--57, 2007.

\bibitem[Owoyele and Pal(2021)]{owoyele2021novel}
Opeoluwa Owoyele and Pinaki Pal.
\newblock A novel machine learning-based optimization algorithm (activo) for
  accelerating simulation-driven engine design.
\newblock \emph{Applied Energy}, 285:\penalty0 116455, 2021.

\bibitem[Shi and Eberhart(1998)]{shi1998parameter}
Yuhui Shi and Russell~C Eberhart.
\newblock Parameter selection in particle swarm optimization.
\newblock In \emph{International conference on evolutionary programming}, pages
  591--600. Springer, 1998.

\bibitem[Trelea(2003)]{trelea2003particle}
Ioan~Cristian Trelea.
\newblock The particle swarm optimization algorithm: convergence analysis and
  parameter selection.
\newblock \emph{Information processing letters}, 85\penalty0 (6):\penalty0
  317--325, 2003.

\bibitem[Sutton et~al.(1998)Sutton, Barto, et~al.]{sutton1998introduction}
Richard~S Sutton, Andrew~G Barto, et~al.
\newblock Introduction to reinforcement learning.
\newblock 1998.

\bibitem[Xu and Pi(2020)]{xu2020reinforcement}
Yue Xu and Dechang Pi.
\newblock A reinforcement learning-based communication topology in particle
  swarm optimization.
\newblock \emph{Neural Computing and Applications}, 32\penalty0 (14):\penalty0
  10007--10032, 2020.

\bibitem[Li and Malik(2016)]{li2016learning}
Ke~Li and Jitendra Malik.
\newblock Learning to optimize.
\newblock \emph{arXiv preprint arXiv:1606.01885}, 2016.

\bibitem[Samma et~al.(2016)Samma, Lim, and Saleh]{samma2016new}
Hussein Samma, Chee~Peng Lim, and Junita~Mohamad Saleh.
\newblock A new reinforcement learning-based memetic particle swarm optimizer.
\newblock \emph{Applied Soft Computing}, 43:\penalty0 276--297, 2016.

\bibitem[Schulman et~al.(2017)Schulman, Wolski, Dhariwal, Radford, and
  Klimov]{schulman2017proximal}
John Schulman, Filip Wolski, Prafulla Dhariwal, Alec Radford, and Oleg Klimov.
\newblock Proximal policy optimization algorithms.
\newblock \emph{arXiv preprint arXiv:1707.06347}, 2017.

\bibitem[Huang and Wang(2020)]{huang2020deep}
Bin Huang and Jianhui Wang.
\newblock Deep-reinforcement-learning-based capacity scheduling for pv-battery
  storage system.
\newblock \emph{IEEE Transactions on Smart Grid}, 12\penalty0 (3):\penalty0
  2272--2283, 2020.

\bibitem[Yu et~al.(2021)Yu, Velu, Vinitsky, Wang, Bayen, and
  Wu]{yu2021surprising}
Chao Yu, Akash Velu, Eugene Vinitsky, Yu~Wang, Alexandre Bayen, and Yi~Wu.
\newblock The surprising effectiveness of ppo in cooperative, multi-agent
  games.
\newblock \emph{arXiv preprint arXiv:2103.01955}, 2021.

\bibitem[Brockman et~al.(2016)Brockman, Cheung, Pettersson, Schneider,
  Schulman, Tang, and Zaremba]{brockman2016openai}
Greg Brockman, Vicki Cheung, Ludwig Pettersson, Jonas Schneider, John Schulman,
  Jie Tang, and Wojciech Zaremba.
\newblock Openai gym.
\newblock \emph{arXiv preprint arXiv:1606.01540}, 2016.

\bibitem[Storn and Price(1997)]{storn1997differential}
Rainer Storn and Kenneth Price.
\newblock Differential evolution--a simple and efficient heuristic for global
  optimization over continuous spaces.
\newblock \emph{Journal of global optimization}, 11\penalty0 (4):\penalty0
  341--359, 1997.

\bibitem[Virtanen et~al.(2020)Virtanen, Gommers, Oliphant, Haberland, Reddy,
  Cournapeau, Burovski, Peterson, Weckesser, Bright, et~al.]{virtanen2020scipy}
Pauli Virtanen, Ralf Gommers, Travis~E Oliphant, Matt Haberland, Tyler Reddy,
  David Cournapeau, Evgeni Burovski, Pearu Peterson, Warren Weckesser, Jonathan
  Bright, et~al.
\newblock Scipy 1.0: fundamental algorithms for scientific computing in python.
\newblock \emph{Nature methods}, 17\penalty0 (3):\penalty0 261--272, 2020.

\bibitem[Mebane~Jr and Sekhon(1997)]{mebane1997genetic}
Walter~R Mebane~Jr and Jasjeet~S Sekhon.
\newblock Genetic optimization using derivatives (genoud).
\newblock \emph{Computer program available upon request}, 1997.

\bibitem[Ali et~al.(2005)Ali, Khompatraporn, and Zabinsky]{ali2005numerical}
M~Montaz Ali, Charoenchai Khompatraporn, and Zelda~B Zabinsky.
\newblock A numerical evaluation of several stochastic algorithms on selected
  continuous global optimization test problems.
\newblock \emph{Journal of global optimization}, 31\penalty0 (4):\penalty0
  635--672, 2005.

\bibitem[gbhat(2021)]{gbhat}
gbhat.
\newblock Particle swarm optimization on non-convex function.
\newblock 2021.

\bibitem[Hedar(2007)]{hedar2007global}
Abdel-Rahman Hedar.
\newblock Global optimization test problems.
\newblock \emph{Disponible Online en: http://wwwoptima. amp. i. kyoto-u. ac.
  jp/member/student/hedar/Hedar\_files/TestGO. htm}, page~43, 2007.

\bibitem[Molga and Smutnicki(2005)]{molga2005test}
Marcin Molga and Czes{\l}aw Smutnicki.
\newblock Test functions for optimization needs.
\newblock \emph{Test functions for optimization needs}, 101:\penalty0 48, 2005.

\end{thebibliography}

\end{document}